\documentclass[10pt,twocolumn,letterpaper]{article}

\usepackage{iccv}
\usepackage{times}
\usepackage{epsfig}
\usepackage{graphicx}
\usepackage{amsmath}
\usepackage{amssymb}
\usepackage{hhline}
\usepackage{multirow}
\usepackage{balance}
\usepackage{pifont}
\usepackage{subfigure}
\makeatletter
\newcommand*\bigcdot{\mathpalette\bigcdot@{.5}}
\newcommand*\bigcdot@[2]{\mathbin{\vcenter{\hbox{\scalebox{#2}{$\m@th#1\bullet$}}}}}
\makeatother
\newcommand{\norm}[1]{\left\lVert#1\right\rVert}

\usepackage[breaklinks=true,bookmarks=false]{hyperref}

\iccvfinalcopy 

\def\iccvPaperID{2282} 

\ificcvfinal\fi
\begin{document}


\title{ViSiL: Fine-grained Spatio-Temporal Video Similarity Learning} %

\author{Giorgos Kordopatis-Zilos$^{1,2}$, Symeon Papadopoulos$^1$, Ioannis Patras$^2$, Ioannis Kompatsiaris$^1$\\
$^1$Information Technologies Institute, CERTH, Thessaloniki, Greece \\
$^2$Queen Mary University of London, Mile End road, E1 4NS London, UK \\
{\tt\small \{georgekordopatis,papadop,ikom\}@iti.gr \hspace*{1em} i.patras@qmul.ac.uk}}

\maketitle

\begin{abstract}
In this paper we introduce ViSiL, a Video Similarity Learning architecture that considers fine-grained Spatio-Temporal relations between pairs of videos -- such relations are typically lost in previous video retrieval approaches that embed the whole frame or even the whole video into a vector descriptor before the similarity estimation. By contrast, our Convolutional Neural Network (CNN)-based approach is trained to calculate video-to-video similarity from refined frame-to-frame similarity matrices, so as to consider both intra- and inter-frame relations. In the proposed method, pairwise frame similarity is estimated by applying Tensor Dot (TD) followed by Chamfer Similarity (CS) on regional CNN frame features - this avoids feature aggregation before the similarity calculation between frames. Subsequently, the similarity matrix between all video frames is fed to a four-layer CNN, and then summarized using Chamfer Similarity (CS) into a video-to-video similarity score -- this avoids feature aggregation before the similarity calculation between videos and captures the temporal similarity patterns between matching frame sequences. We train the proposed network using a triplet loss scheme and evaluate it on five public benchmark datasets on four different video retrieval problems where we demonstrate large improvements in comparison to the state of the art. The implementation of ViSiL is publicly available\footnote{https://github.com/MKLab-ITI/visil}.
\end{abstract}

\section{Introduction}

\begin{figure}[t]
\centering
\includegraphics[width=8.2cm]{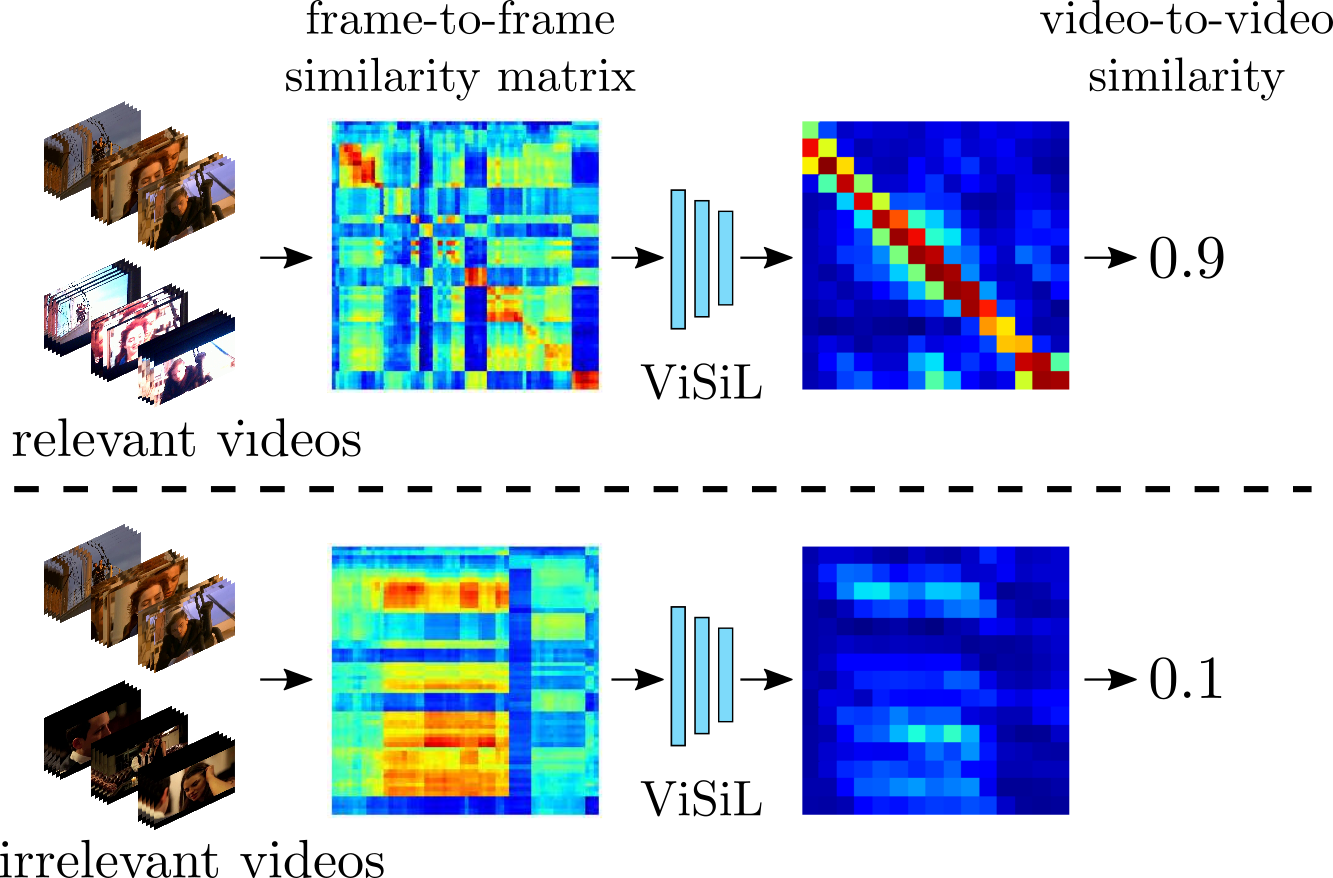}
\vspace{0.05cm}
\caption{Depiction of the frame-to-frame similarity matrix and the CNN output of the ViSiL approach for two video pair examples: relevant videos that contain footage from the same incident (top), unrelated videos with spurious visual similarities (bottom).}
\label{fig:similarity_matrix_example}
\end{figure}

Due to the popularity of Internet-based video sharing services, the volume of video content on the Web has reached unprecedented scales. 
For instance, YouTube reports almost two billion users and more than one billion hours of video viewed per day\footnote{https://www.youtube.com/yt/about/press/, accessed 21 March 2019}.
As a result, content-based video retrieval, which is an essential component in applications such as video filtering, recommendation, copyright protection and verification, becomes increasingly challenging.

In this paper, we address the problem of similarity estimation between pairs of videos, an issue that is central to several video retrieval systems. A straightforward approach to this is to aggregate/pool frame-level features into a single video-level representation on which subsequently one can calculate a similarity measure. Such video-level representations include global vectors \cite{wu2007, gao2017, kordopatis2017b}, hash codes \cite{song2011, liong2017, song2018} and Bag-of-Words (BoW) \cite{cai2011, kordopatis2017a, liao2018}. However, this disregards the spatial and the temporal structure of the visual similarity, as aggregation of features is influenced by clutter and irrelevant content. Other approaches attempt to take into account the temporal sequence of frames in the similarity computation, e.g., by using Dynamic Programming \cite{chou2015, liu2017}, Temporal Networks \cite{tan2009, jiang2016} and Temporal Hough Voting \cite{douze2010, jiang2014}. Another line of research considers spatio-temporal video representation and matching based on Recurrent Neural Networks (RNN) \cite{feng2018, hu2018} or in the Fourier domain \cite{revaud2013, poullot2015, baraldi2018}. Such approaches may achieve high performance in certain tasks such as video alignment or copy detection, but not in more general retrieval tasks. 



A promising direction is exploiting better the spatial and temporal structure of videos in the similarity calculation \cite{douze2010, jiang2014, jiang2016}. However, recent approaches either focused on the spatial processing of frames and completely disregarded temporal information \cite{gao2017, kordopatis2017b}, or considered global frame representations (essentially discarding spatial information) and then considered the temporal alignment among such frame representations \cite{chou2015,baraldi2018}. 
In this paper, we propose ViSiL, a video similarity learning network that considers both the spatial (intra-frame) and temporal (inter-frame) structure of the visual similarity. We first introduce a frame-to-frame similarity that employs Tensor Dot (TD) product and Chamfer Similarity (CS) on {\it region-level} frame Convolutional Neural Network (CNN) features weighted with an attention mechanism. This leads to a frame-to-frame similarity function that takes into consideration region-to-region pairwise similarities, instead of calculating the similarity of frame-level embeddings where the regional details are lost. Then, we calculate the matrix with the similarity scores between each pair of frames between the two videos and use it as input to a four-layer CNN, that is followed by a Chamfer Similarity (i.e., a mean-max filter) at its final layer. By doing so, we learn the temporal structure of the frame-level similarity of relevant videos, such as the presence of diagonal structures in Figure \ref{fig:similarity_matrix_example}, and suppress spurious pairwise frame similarities that might occur.



We evaluate ViSiL on several video retrieval problems, namely Near-Duplicate Video Retrieval (NDVR), Fine-grained Incident and Event-based Video Retrieval (FIVR, EVR), and Action Video Retrieval (AVR) using public benchmark datasets, where in all cases, often by a large margin, it outperforms the state-of-the-art.


\section{Related Work}

Video retrieval approaches can be roughly classified into three categories \cite{liu2013}, namely, methods that calculate similarity using global video representations, methods that account for similarities between individual video frames and methods that employ spatio-temporal video representations.

Methods in the first category extract a global video vector and use dot product or Euclidean distance to compute similarity between videos. 
Goa et al. \cite{gao2017} extracted a \textit{video imprint} for the entire video based on a feature alignment procedure that exploits the temporal correlations and removes feature redundancies across frames. 
Kordopatis et al. created visual codebooks for features extracted from intermediate CNN layers \cite{kordopatis2017a} and employed Deep Metric Learning (DML) to train a network using a triplet loss scheme 
to learn an embedding that minimizes the distance between related videos and maximizes it between irrelevant ones \cite{kordopatis2017b}. A popular direction is the generation of a hash code for the entire video combined with Hamming distance. Liong et al. \cite{liong2017} employed a CNN architecture to learn binary codes for the entire video and trained it end-to-end based on the pair-wise distance of the generated codes and video class labels. Song et al. \cite{song2018} built a self-supervised video hashing system, able to capture the temporal relation between frames using an encoder-decoder scheme. These methods are typically outperformed by the ones of the other two categories. 

\begin{figure*}[t]
\centering
\includegraphics[width=16cm]{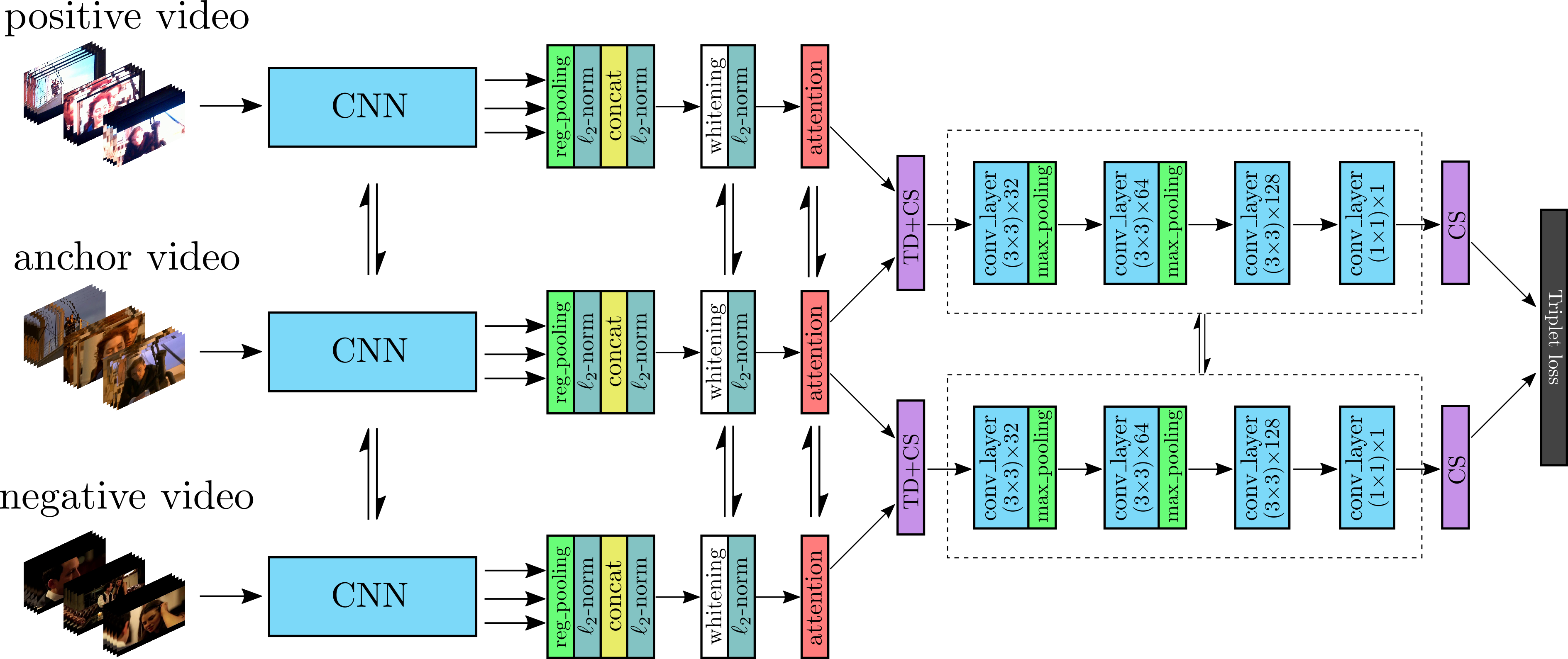}
\vspace{0.1cm}
\caption{Overview of the training scheme of the proposed architecture. A triplet of an anchor, positive and negative videos is provided to a CNN to extract regional features that are PCA whitened and weighted based on an attention mechanism. Then the Tensor Dot product is calculated for the anchor-positive and anchor-negative pairs followed by Chamfer Similarity to generate frame-to-frame similarity matrices. The output matrices are passed to a CNN to capture temporal relations between videos and calculate video-to-video similarity by applying Chamfer Similarity on the output. The network is trained with the triplet loss function. The double arrows indicate shared weights.} 
\label{fig:approach_overview}
\end{figure*}

Methods in the second category typically extract frame-level features to apply frame-to-frame similarity calculation and then aggregate them into video-level similarities. 
Tan et al. \cite{tan2009} proposed a graph-based Temporal Network (TN) structure generated through keypoint frame matching, which is used for the detection of the longest shared path between two compared videos. Several recent works have employed modifications of this approach for the problem of partial-copy detection, combining it with global CNN features \cite{jiang2016} and a CNN+RNN architecture \cite{hu2018}. Additionally, 
other approaches employ
Temporal Hough Voting \cite{douze2010, jiang2014} to align matched frames by means of a temporal Hough transform. These are often outperformed by TN in several related problems. Another popular solution is based on Dynamic Programming (DP) \cite{chou2015, liu2017}. Such works calculate the similarity matrix between all frame pairs, and then extract the diagonal blocks with the largest similarity. To increase flexibility, they also allow limited horizontal and vertical movements. Chou et al. \cite{chou2015} and Liu et al. \cite{liu2017} combined DP with BoW matching to measure frame similarities. However, the proposed solutions are not capable of capturing a large variety of temporal similarity patterns
due to their rigid aggregation approach. By contrast, ViSiL, which belongs to this category of methods, learns the similarity patterns in the CNN subnet that operates on the similarity matrix between the frame pairs.


Methods in the third category extract spatio-temporal representations based on frame-level features and use them to calculate video similarity. A popular direction is to use  the Fourier transform 
in a way that accounts for the temporal structure of video similarity. 
Revaud et al. \cite{revaud2013} proposed the Circulant Temporal Encoding (CTE) that encodes the frame features in a spatio-temporal representation with Fourier transform and thus compares videos in the frequency domain. Poullot et al. \cite{poullot2015} introduced the Temporal Matching Kernel (TMK) that encodes sequences of frames with periodic kernels that take into account the frame descriptor and timestamp. 
Baraldi et al. \cite{baraldi2018} built a deep learning layer component based on TMK and set up a training process to learn the feature transform coefficients using a triplet loss that takes into account both the video similarity score and the temporal alignment. However, the previous methods rely on global frame representations, which disregard the spatial structure of similarity. 
Finally, Feng et al. \cite{feng2018} developed an approach based on cross gated bilinear matching for video re-localization. They employed C3D features \cite{tran2015} and built a multi-layer recurrent architecture that matches videos through attention weighting and factorized bilinear matching to locate related video parts. However, even though this approach performs well on video matching problems, it was found to be inapplicable for video retrieval tasks as will be shown in Section \ref{sec:experiments}. 


\section{Preliminaries}
\label{sec:preliminaries}

\textit{Tensor Dot} (\textbf{TD}): Having two tensors $\mathcal{A} \in \mathbb{R}^{N_1 \times N_2 \times K}$ and $\mathcal{B} \in \mathbb{R}^{K \times M_1 \times M_2}$, their TD (also known as tensor contraction) is given by summing the two tensors over specific axes. Following the notation in \cite{yang2017}, TD of two tensors is
\begin{equation}
\mathcal{C} = \mathcal{A} \bigcdot _{(i,j)} \mathcal{B}
\label{eq:tensor_dot}
\end{equation}
where $\mathcal{C}\in \mathbb{R}^{N_1 \times N_2 \times M_1 \times M_2}$ is the TD of the tensors, and $i$ and $j$ indicate the axes over which the tensors 
are summed. In the given example $i$ and $j$ can only be $3$ and $1$ respectively, since they are the only ones of the same size ($K$). 

\textit{Chamfer Similarity} (\textbf{CS}): This is the similarity counterpart of Chamfer Distance \cite{barrow1977}. Considering two sets of items $x$ and $y$ with total number of $N$ and $M$ items respectively and their similarity matrix $\mathcal{S}\in \mathbb{R}^{N \times M}$, CS is calculated as the average similarity of the most similar item in set $y$ for each item in set $x$. 
This is formulated in Equation \ref{eq:chamfer_similarity}.
\begin{equation}
\text{CS}(x, y) = \frac{1}{N} \sum_{i=1}^{N} \, \max_{j \in [1,M]}  \mathcal{S}(i,j)
\label{eq:chamfer_similarity}
\end{equation}
Note that CS is not symmetric, i.e. $\text{CS}(x,y)\neq \text{CS}(y, x)$, however, that a symmetric variant SCS can be defined as, $\text{SCS}(x, y) = (\text{CS}(x, y) + \text{CS}(y, x))/2$.


\section{ViSiL description}

Figure \ref{fig:approach_overview} illustrates the proposed approach. We first extract features from the intermediate convolution layers of a CNN architecture by applying region pooling on the feature maps. These are further PCA-whitened and weighted based on an attention mechanism (section \ref{sec:feature_extraction}). Additionally, a similarity function based on TD and CS is devised to accurately compute the similarity between frames (section \ref{sec:frame_level_similarity}). A similarity matrix comprising all pairwise frame similarities is then fed to a CNN to train a video-level similarity model (section \ref{sec:video_level_similarity}). 
This is trained with a triplet loss scheme (section \ref{sec:loss_function}) based on selected and automatically generated triplets from a training dataset (section \ref{sec:training_process}). 

\subsection{Feature extraction}
\label{sec:feature_extraction}

Given an input video frame, we apply Regional Maximum Activation of Convolution (R-MAC) \cite{tolias2015} on the activations of the intermediate convolutional layers \cite{kordopatis2017a} given a specific granularity level $L_N, N\in \{1,2,3,...\}$. Given a CNN architecture with a total number of $K$ convolutional layers, this process generates $K$ feature maps $\mathcal{M}^k \in \mathbb{R}^{N \times N \times C_k} (k=1, ..., K)$, where $C_k$ is the number of channels of the $k^{th}$ convolution layer. All extracted feature maps have the same resolution ($N \times N$) and are concatenated into a frame representation $\mathcal{M} \in \mathbb{R}^{N \times N \times C}$, where $C = C_1+...+C_K$. We also apply $\ell^2$-normalization on the channel axis of the feature maps, before and after  concatenation. This feature extraction process is denoted as L$_N$-iMAC. The extracted frame features retain the spatial information of frames at different granularities. 
We then employ PCA on the extracted frame descriptors to perform whitening and/or dimensionality reduction as in \cite{jegou2012}. 


$\ell^2$-normalization on the extracted frame descriptors result in all region vectors being equally considered in the similarity calculation. For instance, this would mean that a completely dark region would have the same impact on similarity with a region depicting a subject of interest. To avoid this issue, we weight the frame regions based on their saliency via a visual attention mechanism over region vectors inspired by methods from different research fields, i.e. document classification \cite{yang2016}. 
To successfully adapt it to the needs of video retrieval, we build the following attention mechanism: given a frame representation $\mathcal{M}$ with region vector $\textbf{r}_{ij}: \mathcal{M}(i,j,\cdot) \in \mathbb{R}^{C}$, where $i\in[1,N], j\in[1,N]$, we introduce a visual context unit vector $\textbf{u}$ and use it to measure the importance of each region vector. To this end, we calculate the dot product between every $\textbf{r}_{ij}$ region vector, with the internal context vector $\textbf{u}$ to derive the weight scores $\alpha_{ij}$. Since all vectors are unit norm, $\alpha_{ij}$ will be in the range $[-1,1]$. To retain region vectors' direction and change their norm, we divide the weight scores $\alpha_{ij}$ by 2 and add 0.5 in order to be in range $[0,1]$. Equation \ref{eq:attention_mechanism} formulates the weighting process.
\begin{equation}
    \begin{aligned}
    & \alpha_{ij} = \textbf{u}^\top \textbf{r}_{ij}, \quad s.t. \norm{\textbf{u}} = 1 \\
    & \textbf{r}_{ij}' = (\alpha_{ij}/2 + 0.5)\textbf{r}_{ij}
    \end{aligned}
    \label{eq:attention_mechanism}
\end{equation}


All functions in the weighting process are differentiable; therefore $\textbf{u}$ is learned through the training process. Unlike the common practice in the literature, we do not apply any normalization function on the calculated weights (e.g. softmax or division by sum) because we want to weight each vector independently. Also, we empirically found that, unlike other works, using a hidden layer in the attention module has negative effect on the system's performance.

\begin{figure}[t]
\centering
\includegraphics[width=7.5cm]{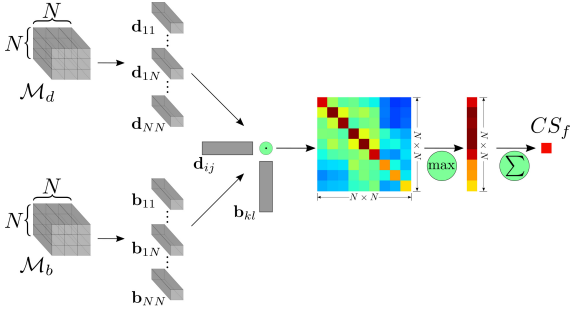}
\caption{Illustration of frame-level similarity calculation between two video frames. In this example, the frames are near duplicates.}
\label{fig:frame_level_similarity}
\end{figure}

\subsection{Frame-to-frame similarity}
\label{sec:frame_level_similarity}

Given two video frames $d$, $b$, we apply CS on their region feature maps to calculate their similarity (Figure \ref{fig:frame_level_similarity}). 
First, the regional feature maps $\mathcal{M}_d, \mathcal{M}_b \in \mathbb{R}^{N \times N \times C}$ 
are decomposed into their region vectors $\textbf{d}_{ij},\textbf{b}_{kl} \in \mathbb{R}^{C}$. Then, the dot product between every pair of region vectors is calculated, creating the similarity matrix of the two frames, and CS is applied on the similarity matrix to compute the frame-to-frame similarity.
\begin{equation}
    \begin{aligned}
    \text{CS}_f(d, b) = \frac{1}{N^2}\sum_{i,j=1}^{N} \max_{k, l \in [1,N]} \textbf{d}_{ij}^\top\textbf{b}_{kl}
    \end{aligned}
    \label{eq:frame_level_similarity}
\end{equation}
This process leverages the geometric information captured by region vectors and provides some degree of spatial invariance. More specifically, the CNN extracts features that correspond to mid-level visual structures, such as object parts, and combined with CS, that by design disregards the global structure of the region-to-region matrix, constitutes a robust similarity calculation process against spatial transformations, e.g. spatial shift. This presents a trade-off between the preservation of the frame structure and invariance to spatial transformations.


\subsection{Video-to-video similarity}
\label{sec:video_level_similarity}

To apply frame-to-frame similarity on two videos $q$, $p$ with $X$ and $Y$ frames respectively,  we apply TD combined with CS on the corresponding video tensors $\mathcal{Q}$ and $\mathcal{P}$ and derive the frame-to-frame similarity matrix $\mathcal{S}_f^{qp}\in \mathbb{R}^{X\times Y}$. This is formulated in Equation \ref{eq:similarity_matrix}.
\begin{equation}
    \begin{aligned}
    \mathcal{S}_f^{qp} = \frac{1}{N^2}\sum_{i=1}^{N^2} \max_{j \in [1,N^2]} \mathcal{Q} \bigcdot _{(3,1)} \mathcal{P}^\top (\cdot, i, j, \cdot)
    \end{aligned}
    \label{eq:similarity_matrix}
\end{equation}
where the TD axes indicate the channel dimension of the corresponding video tensors. In that way, we apply Equation \ref{eq:frame_level_similarity} on every frame pair.

\begin{table}[t]
  \centering
  \scalebox{0.9}{
  \begin{tabular}{|l|c|c|c|}
    \hline
      \multirow{2}{*}{\textbf{Type}}        &   \textbf{Kernel size}   &   \multirow{2}{*}{\textbf{Output size}} & \multirow{2}{*}{\textbf{Activ.}}   \\ 
      &  \textbf{/ stride} & & \\ \hline\hline
      \textbf{Conv}       &  3$\times$3 / 1   &  $X \times Y \times$ 32  & ReLU           \\ \hline
      \textbf{M-Pool}     &  2$\times$2 / 2   &  $X$/2 $\times Y$/2 $\times$ 32  & ---    \\ \hline
      \textbf{Conv}       &  3$\times$3 / 1   &  $X$/2 $\times Y$/2 $\times$ 64  & ReLU   \\ \hline
      \textbf{M-Pool}     &  2$\times$2 / 2   &  $X$/4 $\times Y$/4 $\times$ 64  & ---    \\ \hline
      \textbf{Conv}       &  3$\times$3 / 1   &  $X$/4 $\times Y$/4 $\times$ 128 & ReLU   \\ \hline
      \textbf{Conv}       &  1$\times$1 / 1   &  $X$/4 $\times Y$/4 $\times$ 1   & ---    \\ \hline
    \end{tabular}
    }
  \caption{Architecture of the proposed network for video similarity learning. For the calculation of the output size, we assume that two videos with total number of $X$ and $Y$ frames are provided.}
  \label{tab:cnn_architecture}
\end{table}

To calculate the similarity between two videos, the generated similarity matrix $\mathcal{S}_f^{qp}$ derived from the previous process is provided to a CNN network. The network is capable of learning robust patterns of within-video similarities at segment level. 
Table \ref{tab:cnn_architecture} displays the architecture of the CNN architecture of the proposed ViSiL framework. 

To calculate the final video similarity, we apply the \textit{hard tanh} activation function on the values of the network output, which clips values within range $[-1,1]$. 
Then, we apply CS to derive a single value as in Equation \ref{eq:video_level_similarity}. 
\begin{equation}
	\text{CS}_v(q, p) = \frac{1}{X'} \sum_{i=1}^{X'} \max_{j \in [1,Y']} \text{Htanh}(\mathcal{S}_v^{qp}(i, j))
	\label{eq:video_level_similarity}
\end{equation}
where $\mathcal{S}_v^{qp}\in\mathbb{R}^{X'\times Y'}$ is the output of the CNN network, and Htanh indicates the element-wise hard tanh function. The output of the network has to be bounded in order to accordingly set the margin in Equation \ref{eq:triplet_loss}. 

Similar to the frame-to-frame similarity calculation, this process is a trade-off between  respecting video-level structure and being invariant to some temporal differences. As a result, different temporal similarity structures in the frame-to-frame similarity matrix can be captured, e.g. strong diagonals or diagonal parts (i.e. contained sequences).

\subsection{Loss function}
\label{sec:loss_function}
The target video similarity score CS$_v(q, p)$ should be higher for relevant videos and lower for irrelevant ones. To train our network we organize our video collection in video triplets ($v$, $v^+$, $v^-$), where $v$, $v^+$, $v^-$ stand for an anchor, a positive (i.e. relevant), and a negative (i.e. irrelevant) video respectively. To force the network to assign higher similarity scores to positive video pairs and lower to negative ones, we use the `triplet loss', that is 
\begin{equation}
    \mathcal{L}_{tr} = \max\{0, \text{CS}_v(v, v^-) - \text{CS}_v(v, v^+) + \gamma\}
    \label{eq:triplet_loss}
\end{equation}
where $\gamma$ is a margin parameter. 


In addition, we define a similarity regularization function that penalizes high values in the input of hard tanh that would lead to saturated outputs. This is an effective mechanism to drive the network to generate output matrices $\mathcal{S}_v$ with values in the range $[-1, 1]$, which is the clipping range of hard tanh. To calculate the regularization loss, we simply sum all values in the output similarity matrices that fall outside the clipping range 
(Equation \ref{eq:similarity_loss}).

\begin{equation}
    \begin{aligned}
        \mathcal{L}_{reg} = \sum_{i=1}^{X'}\sum_{j=1}^{Y'} |\max\{0, \mathcal{S}_v^{qp}(i,j)-1\}|  +\\
          + |\min\{0, \mathcal{S}_v^{qp}(i,j)+1\}|
    \end{aligned}
    \label{eq:similarity_loss}
\end{equation}

Finally, the total loss function is given in Equation \ref{eq:total_loss}.
\begin{equation}
    \begin{aligned}
        \mathcal{L} = \mathcal{L}_{tr} + r * \mathcal{L}_{reg}
    \end{aligned}
    \label{eq:total_loss}
\end{equation}
where $r$ is a regularization hyperparameter that tunes the contribution of the similarity regularization to the total loss.

\subsection{Training ViSiL}
\label{sec:training_process}

Training the ViSiL architecture requires a training dataset with ground truth annotations at segment level. 
Using such annotations, we extract video pairs with related visual content to serve as anchor-positive pairs during training. Additionally, we artificially generate positive videos by applying a number of transformations on arbitrary videos. 
We consider three categories of transformation: (i) \textit{colour}, including conversion to grayscale, brightness, contrast, hue, and saturation adjustment, (ii) \textit{geometric}, including horizontal or vertical flip, crop, rotation, resize and rescale, and (iii) \textit{temporal}, including slow motion, fast forward, frame insertion, video pause or reversion. During training, one transformation from each category is randomly selected and applied on the selected video.



We construct two video pools that consist of positive pairs. For each positive pair we then generate \textit{hard triplets}, i.e. construct negative videos (hard negatives) with similarity to the anchor that is greater than the one between the anchor and positive videos. In what follows, we use a BoW approach \cite{kordopatis2017a} to calculate similarities between videos. 

The first pool derives from the annotated videos in the training dataset. Two videos with at least five second overlap constitute a positive pair. Let $s$ be the similarity of the corresponding video segments. Videos with similarity (BoW-based \cite{kordopatis2017a}) larger than $s$ with either of the segments in the positive pair, constitute hard negatives. The second pool derives from arbitrary videos from the training dataset that are used to artificially generate positive pairs. Videos that are similar with the initial videos (similarity $>0.1$) are considered  hard negatives. To avoid potential near-duplicates, we exclude videos with similarity $>0.5$ from the hard negative sets.

At each training epoch, we sample $T$ triplets from each video pool. Due to GPU memory limitations, we do not feed the entire videos to the network. Instead, we select a random video snippet with total size of $W$ frames from each video in the triplet, assuring that there are at least five seconds overlap between the anchor and the positive videos.

\section{Evaluation setup}
\label{sec:experiments}

The proposed approach is evaluated on four retrieval tasks, namely Near-Duplicate Video Retrieval (NDVR), Fine-grained Incident Video Retrieval (FIVR), Event Video Retrieval (EVR), and Action Video Retrieval (AVR). 
In all cases, we report the mean Average Precision (mAP).

\subsection{Datasets}
\label{sec:datasets}

\textbf{VCDB} \cite{jiang2014} is used as the training dataset to generate triplets for training our models. It consists of 528 videos with 9,000 pairs of copied segments in the core dataset, and also a subset of 100,000 distractor videos.


\textbf{CC\_WEB\_VIDEO} \cite{wu2007} simulates the NDVR problem. 
It consists of 24 query sets and 13,129 videos. We found several quality issues with the annotations, e.g. numerous positives mislabeled as negatives. 
Hence, we provide results on a `cleaned' version of the annotations. We also use two evaluation settings, one measuring performance only on the query sets, and a second  on the entire dataset. 

\textbf{FIVR-200K} is used for the FIVR task \cite{kordopatis2018}. 
It consists of 225,960 videos and 100 queries. 
It includes three different retrieval tasks: a) the Duplicate Scene Video Retrieval (DSVR), b) the Complementary Scene Video Retrieval (CSVR), and c) the Incident Scene Video Retrieval (ISVR).
For quick comparison of the different variants, we use \textbf{FIVR-5K}, a subset of FIVR-200K by selecting the 50 most difficult queries in the DSVR task (using \cite{kordopatis2017a} to measure difficulty), and for each one  randomly picking the 30\% of annotated videos per label category. 


\textbf{EVVE} \cite{revaud2013} was designed for the EVR problem. It consists of 2,375 videos, and 620 queries. However, we managed to download and process only 1897 videos and 503 queries ($\approx$80\% of the initial dataset) due to the unavailability of the remaining ones. 

Finally, \textbf{ActivityNet} \cite{caba2015}, reorganized based on \cite{feng2018}, is used for the AVR task. It consists of 3,791 training, 444 validation and 494 test videos. The annotations contain the exact video segments that correspond to specific actions. For evaluation, we consider any pair of videos with at least one common label as related. 



\subsection{Implementation details}
\label{sec:experiments}

We extract one frame per second for each video. 
For all retrieval problems except for AVR, we are using the feature extraction scheme of Section \ref{sec:feature_extraction} based on ResNet-50 \cite{he2016},
but for efficiency purposes only extract intermediate features from the output maps of the four residual blocks. Additionally, the PCA for the whitening layer is learned from 1M region vectors sampled from videos in VCDB. For AVR, we extract features from the last 3D convolutional layer of the I3D architecture \cite{carreira2017} by max-pooling on the spatial dimensions. We also tested I3D features for the other retrieval problems, but without any significant improvements.

For training, we feed the network with only one video triplet at a time due to GPU memory limitations. 
We employ Adam optimization \cite{kingma2014} with learning rate $l=10^{-5}$. For each epoch, $T$=1000 triplets are selected per pool. The model is trained for 100 epochs, i.e. 200K iterations, and the best network is selected based on mean Average Precision (mAP) on a validation set. Other parameters are set to $\gamma = 0.5$, $r=0.1$ and $W=64$. 
The weights of the feature extraction CNN and whitening layer remain fixed.

\section{Experiments}
\label{sec:experiments}

In this section, we first compare the proposed frame-to-frame similarity calculation scheme 
with several global features with dot product similarity (Section \ref{sec:feature_comparison}). We also provide an ablation study to evaluate the proposed approach under different configurations (Section \ref{sec:ablation_study}). Finally, we compare the ``full'' proposed approach (denoted as ViSiL$_v$) with the best performing methods in the state-of-the-art (to the best of our knowledge) in each problem (Section \ref{sec:soa_comparison}). 
We have re-implemented two popular approaches that employ similarity calculation on frame-level representations, i.e. DP \cite{chou2015} and TN \cite{tan2009}. However, both of them were originally proposed in combination with 
hand-crafted features, which is an outdated practice. 
Hence, we combine them with the proposed feature extraction scheme and our frame-to-frame similarity calculation. 
We also implemented a naive adaptation of the publicly available  Video re-localization (VReL) method \cite{feng2018} to a retrieval setting, where we rank videos based on the probability of the predicted segment (Equation 12 in the original paper).


\subsection{Frame-to-frame similarity comparison}
\label{sec:feature_comparison}
This section presents a comparison  on FIVR-5K of the proposed feature extraction scheme 
against several global pooling schemes proposed in the literature. 
Dot product is used for similarity calculation. 
Video-level similarity for all runs is calculated with the application of the raw CS on the generated similarity matrices. 
The benchmarked feature extraction methods include the Maximum Activations of Convolutions (MAC) \cite{tolias2015},  Sum-Pooled Convolutional features (SPoC) \cite{babenko2015}, Regional Maximum Activation of Convolutions (R-MAC) \cite{tolias2015}, Generalized Mean (GeM) pooling \cite{radenovic2018} (with initial $p=3$ (cf. Table 1 in \cite{radenovic2018}) and intermediate Maximum Activation of Convolutions (iMAC) \cite{kordopatis2017a}, which is equivalent to the proposed feature extraction for $N=1$. Additionally, we evaluate the proposed scheme with region levels $L_N, N=2,3$, and with two different region vector sizes for each region level. We use PCA to reduce region vectors' size, without applying whitening.

\begin{table}[t]
  \centering
  \begin{tabular}{|l|}
    \hline
      \textbf{Features}  \\ \hline\hline
      \textbf{MAC} \cite{tolias2015}    \\ \hline
      \textbf{SPoC} \cite{babenko2015}   \\ \hline
      \textbf{R-MAC} \cite{tolias2015}   \\ \hline
      \textbf{GeM} \cite{hao2017}    \\ \hline
      \textbf{iMAC} \cite{kordopatis2017a}   \\ \hline\hline
      \textbf{L$_2$-iMAC}  \\ \hline
      \textbf{L$_2$-iMAC}  \\ \hline\hline
      \textbf{L$_3$-iMAC}  \\ \hline
      \textbf{L$_3$-iMAC}  \\ \hline
    \end{tabular}
  \begin{tabular}{|c|}
    \hline
      \textbf{Dims.}  \\ \hline\hline
      2048    \\ \hline
      2048    \\ \hline
      2048    \\ \hline
      2048    \\ \hline
      3840    \\ \hline\hline
      4x3840  \\ \hline
      4x512   \\ \hline\hline
      9x3840  \\ \hline
      9x256   \\ \hline
    \end{tabular}
  \begin{tabular}{|c|c|c|}
    \hline
      \textbf{DSVR}   &  \textbf{CSVR}   &  \textbf{ISVR}       \\ \hline\hline
      0.747  &  0.730  &  0.684      \\ \hline
      0.735  &  0.722  &  0.669      \\ \hline
      0.777  &  0.764  &  0.707      \\ \hline
      0.776  &  0.768  &  0.711      \\ \hline
      0.755  &	0.749  &  0.689      \\ \hline\hline
      0.814  &  0.810  &  0.738      \\ \hline
      0.804  &  0.802  &  0.727      \\ \hline\hline
      \textbf{0.838}  &  \textbf{0.832}  &  \textbf{0.739}      \\ \hline
      0.823  &  0.818  &  0.738      \\ \hline
    \end{tabular}
    \caption{mAP comparison of proposed feature extraction and similarity calculation against state-of-the-art feature descriptors with dot product for similarity calculation on FIVR-5K. Video similarity is computed based on CS on the derived similarity matrix.}
    \label{tab:feature_comparison}
\end{table}

Table \ref{tab:feature_comparison} presents the results of the comparison on FIVR-5K.  The proposed scheme with $N=3$ (L$_3$-iMAC) achieves the best results on all evaluation tasks by a large margin. Furthermore, it is noteworthy that the reduced features achieve competitive performance especially compared with the global descriptors of similar dimensionality. Hence, in settings where there is insufficient storage space, the reduced ViSiL features offer an excellent trade-off between retrieval performance and storage cost.
We also tried to combine the proposed scheme with other pooling schemes, e.g. GeM pooling, but this had no noteworthy impact on the system's performance. 
Next, we will consider the best performing scheme (L$_3$-iMAC without dimensionality reduction) 
as the base frame-to-frame similarity scheme \textbf{ViSiL}$_f$.

\subsection{Ablation study}
\label{sec:ablation_study}

 We first evaluate the impact of each individual module of the architecture on the retrieval performance of ViSiL. Table \ref{tab:component_impact} presents the results of four runs with different configuration settings on FIVR-5K. The attention mechanism in the third run is trained using the main training process. 
 The addition of each component offers additional boost to the performance of the system. The biggest improvement for the DSVR and CSVR tasks, 0.024 and 0.021 of mAP respectively, is due to employing a CNN model for refined video-level similarity calculation in \textbf{ViSiL}$_v$. Also, considerable gains on the ISVR task (0.018 mAP) are due to the application of the attention mechanism.
%
%
We also report results when the Symmetric Chamfer Distance (SCS) is used for both frame-to-frame and video-to-video similarity calculation (\textbf{ViSiL}$_{sym}$). 
Apparently, the non symmetric version of the CS works significantly better in this problem.

\begin{table}[h]
  \centering
  \begin{tabular}{|l|c|c|c|}
    \hline
      \textbf{Task}   &   \textbf{DSVR}   &   \textbf{CSVR}   &   \textbf{ISVR}   \\ \hline\hline
      \textbf{ViSiL}$_f$  &  0.838   &  0.832   &  0.739  \\ \hline
      \textbf{ViSiL}$_f$+\textbf{W}   &  0.844   &  0.837   &  0.750  \\ \hline
      \textbf{ViSiL}$_f$+\textbf{W}+\textbf{A} &  0.856   &  0.848   &  0.768 \\ \hline\hline
      \textbf{ViSiL}$_{sym}$ & 0.830   &  0.823   &  0.731  \\ \hline
      \textbf{ViSiL}$_v$  & \textbf{0.880}  &  \textbf{0.869}  &  \textbf{0.777}      \\ \hline
    \end{tabular}
  \caption{Ablation studies on FIVR-5K. \textbf{W} and \textbf{A} stand for whitening and attention mechanism respectively.}
  \label{tab:component_impact}
\end{table}

Additionally, we evaluate the impact of the similarity regularization loss $\mathcal{L}_{reg}$ of Equation \ref{eq:similarity_loss}. This appears to have notable impact on the retrieval performance of the system. The mAP increases for all three tasks reaching an improvement of more than 0.02 mAP on DSVR and ISVR tasks.

\begin{table}[h]
  \centering
  \begin{tabular}{|c|c|c|c|}
    \hline
      \textbf{$\mathcal{L}_{reg}$} &  \textbf{DSVR}   &   \textbf{CSVR}   &   \textbf{ISVR}     \\ \hline\hline
      \ding{55}   & 0.859   &  0.842   &  0.756  \\ \hline
      \checkmark  & \textbf{0.880}  &  \textbf{0.869}  &  \textbf{0.777}      \\ \hline
    \end{tabular}
  \caption{Impact of similarity regularization on the performance of the proposed method on FIVR-5K.}
  \label{tab:similarity_loss}
\end{table}



In the supplementary material we assess the performance of similarity functions other than CS, the impact of different values of hyperparameters $\gamma$, $W$ and $r$, and the computational complexity of the method.

\subsection{Comparison against state-of-the-art}
\label{sec:soa_comparison}

\subsubsection{Near-duplicate video retrieval}
\label{sec:ndvr}

We first compare the performance of ViSiL against state-of-the-art approaches on several versions of CC\_WEB\_VIDEO \cite{wu2007}. 
The proposed approach is compared with the publicly available implementation of Deep Metric Learning (DML) \cite{kordopatis2017b}, the Circulant Temporal Encoding (CTE) \cite{revaud2013} (we report the results of the original paper)  
and our two re-implementations based on Dynamic Programming (DP) \cite{chou2015} and Temporal Networks (TN) \cite{tan2009}. The ViSiL$_v$ approach achieves the best performance compared to all competing systems in all cases except in the case where the original annotations are used (where CTE performs best). In that case, there were several erroneous annotations as explained above. When tested on the `cleaned' version, ViSiL achieves almost perfect results in both evaluation settings. Moreover, it is noteworthy that our re-implementations of the state-of-the-art methods lead to considerably better results than the ones reported in the original papers, meaning that direct comparison with the originally reported results would be much more favourable for ViSiL. 

\begin{table}[h]
  \centering
  \scalebox{1.}{
  \begin{tabular}{|l|c|c|c|c|}
    \hline 
  	\textbf{Method}  & \textbf{cc\_web} & \textbf{cc\_web}$^*$ & \textbf{cc\_web}$_c$ & \textbf{cc\_web}$^*_c$  \\ \hline\hline
    \textbf{DML} \cite{kordopatis2017b}   & 0.971   & 0.941    & 0.979   & 0.959    \\ \hline
    \textbf{CTE} \cite{revaud2013}        & \textbf{0.996}  & ---  & ---  & ---     \\ \hline\hline
    \textbf{DP} \cite{chou2015}           & 0.975   & 0.958    & 0.990   & 0.982    \\ \hline
    \textbf{TN} \cite{tan2009}            & 0.978   & 0.965    & 0.991   & 0.987    \\  \hline\hline
    \textbf{ViSiL}$_f$                    & 0.984   & 0.969    & 0.993   & 0.987    \\ \hline
    \textbf{ViSiL}$_{sym}$                    & 0.982   & 0.969    & 0.991   & 0.988    \\ \hline
    \textbf{ViSiL}$_v$                    & 0.985 & \textbf{0.971} & \textbf{0.996} & \textbf{0.993}     \\ \hline
  \end{tabular}
  }
  \caption {mAP of three ViSiL setups and SoA methods on four different versions of CC\_WEB\_VIDEO. ($^*$) denotes evaluation on the entire dataset, and  subscript $c$ that the cleaned version of the annotations was used.}
  \label{tab:map_depth} 
\end{table}

\setcounter{table}{6}
\begin{table*}[htb]
  \centering
  \scalebox{0.87}{
  \begin{tabular}{|l|}
    \hline 
  	  \textbf{Method}       \\ \hline\hline
      \textbf{LAMV}\cite{baraldi2018}     \\ \hline
      \textbf{LAMV}+QE \cite{baraldi2018} \\ \hline\hline
      \textbf{ViSiL}$_f$   \\ \hline
      \textbf{ViSiL}$_{sym}$   \\ \hline
      \textbf{ViSiL}$_v$   \\ \hline
  \end{tabular}
  \begin{tabular}{|c|}
    \hline 
  	  \textbf{mAP}  \\ \hline\hline
      0.536 \\ \hline
      0.587 \\ \hline\hline
      0.589 \\ \hline
      0.610 \\ \hline
      \textbf{0.631} \\ \hline
  \end{tabular}
  \begin{tabular}{|c|c|c|c|c|c|c|c|c|c|c|c|c|}
    \hline 
  	  \multicolumn{13}{|c|}{\textbf{per event class}} \\ \hline\hline
      0.715 & 0.383 & 0.158 & 0.461 & 0.387 & 0.277 & 0.247 & 0.138 & 0.222 & 0.273 & 0.273 & 0.908 & 0.691  \\ \hline
      0.837 & 0.500 & 0.126 & \textbf{0.588} & \textbf{0.455} & 0.343 & 0.267 & 0.142 & 0.230 & 0.293 & 0.216 & \textbf{0.950} & 0.776  \\ \hline \hline
      0.889 & 0.570 & 0.169 & 0.432 & 0.345 & 0.393 & 0.297 & 0.181 & 0.479 & 0.564 & 0.369 & 0.885 & 0.799 \\ \hline
      0.864 & 0.704 & \textbf{0.357} & 0.440 & 0.363 & 0.295 & \textbf{0.370} & 0.214 & 0.577 & 0.389 & 0.266 & 0.943 & 0.702 \\ \hline
     \textbf{ 0.918} & \textbf{0.724} & 0.227 & 0.446 & 0.390 & \textbf{0.405} & 0.308 & \textbf{0.223} & \textbf{0.604} & \textbf{0.578} & \textbf{0.399} & 0.916 & \textbf{0.855} \\ \hline
  \end{tabular}
  }
  \caption {mAP comparison of three ViSiL setups with the LAMV \cite{baraldi2018} on EVVE. The ordering of events is the same as in \cite{revaud2013}. Our results are reported on a subset of the videos ($\approx$80\% of the original dataset) due to unavailability of the full original dataset.}
  \label{tab:evve_map}
\end{table*}
\setcounter{table}{5}

\begin{figure}[t]
\centering
\includegraphics[width=8.2cm]{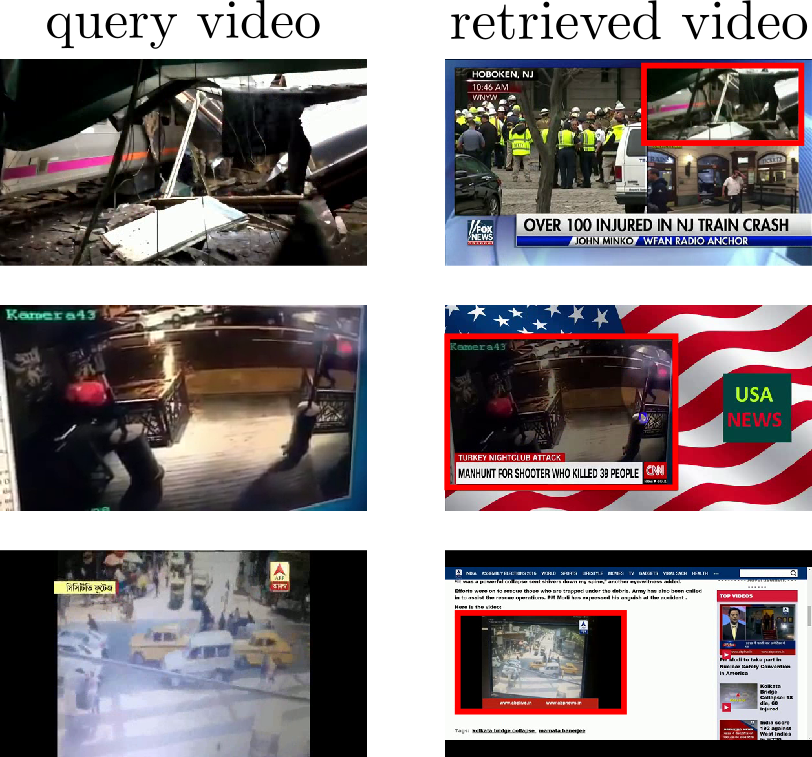}
\caption{Examples of challenging cases of related videos that were mistakenly not labelled as positives in FIVR-200K.}
\label{fig:fail_examples}
\end{figure}

\subsubsection{Fine-grained incident video retrieval}
\label{sec:fivr}

Here, we evaluate the performance of ViSiL against the state-of-the-art approaches on FIVR-200K \cite{kordopatis2018}. We compare with the best performing method reported in the original paper, i.e. Layer Bag-of-Words (LBoW) \cite{kordopatis2017a} implemented with iMAC features from VGG \cite{simonyan2014} and our two re-implementations of DP \cite{chou2015} and TN \cite{tan2009}. Furthermore, we tested our adaptation of VReL \cite{feng2018}, but with no success (neither when training on VCDB nor on ActivityNet). 
As shown in Table \ref{tab:fivr}, ViSiL$_v$ outperforms all competing systems, including DP and TN. Its performance is considerably higher on the DSVR task achieving almost 0.9 mAP. When conducting manual inspection of the erroneous results, we came across some interesting cases (among the top ranked irrelevant videos), which should actually be considered as positive results but were not labelled as such  (Figure \ref{fig:fail_examples}).

\subsubsection{Event video retrieval}
\label{sec:evr}

For EVR, we compare ViSiL with the state-of-the-art approach Learning to Align and Match Videos (LAMV) \cite{baraldi2018}. ViSiL performs well on the EVR problem, even without applying any query expansion technique, i.e. Average Query Expansion (AQE) \cite{douze2013}. As shown in Table \ref{tab:evve_map}, ViSiL$_v$ achieves the best results on the majority of the events in the dataset. However, due to the fact that some of the videos are no longer available, we report results on the currently available ones that account for $\approx$80\% of the original EVVE dataset.

\subsubsection{Action video retrieval}
\label{sec:avr}

We also assess the performance of the proposed approach on ActivityNet \cite{caba2015} reorganized based on \cite{feng2018}. We compare with the publicly available DML approach  \cite{kordopatis2017b}, our re-implementations of DP \cite{chou2015} and TN \cite{tan2009}, and the adapted VReL \cite{feng2018}. For all runs, we extracted features from I3D \cite{carreira2017}. The proposed approach with the symmetric similarity calculation ViSiL$_{sym}$ outperforms all other approaches by a considerable margin (0.035 mAP) to the second best. 

\begin{table}[t]
  \centering
  \scalebox{1.}{
  \begin{tabular}{|l|c|c|c|c|}
    \hline 
  	\textbf{Run}        &  \textbf{DSVR}   & \textbf{CSVR}    & \textbf{ISVR}    \\ \hline\hline
  	\textbf{LBoW} \cite{kordopatis2017a}      & 0.710   & 0.675   & 0.572   \\ \hline\hline
    \textbf{DP} \cite{chou2015}               & 0.775   & 0.740   & 0.632   \\ \hline
    \textbf{TN} \cite{tan2009}                & 0.724   & 0.699   & 0.589   \\ \hline\hline
    \textbf{ViSiL}$_f$                        & 0.843   & 0.797   & 0.660   \\ \hline
    \textbf{ViSiL}$_{sym}$                        & 0.833   & 0.792   & 0.654   \\ \hline
    \textbf{ViSiL}$_v$                        & \textbf{0.892}   & \textbf{0.841}   & \textbf{0.702}       \\ \hline
  \end{tabular}
  }
  \caption {mAP comparison of three ViSiL setups and state-of-the-art methods on the three tasks of FIVR-200K.}
  \label{tab:fivr}
\end{table}

\setcounter{table}{7}
\begin{table}[t]
  \centering
  \scalebox{1.}{
  \begin{tabular}{|l|c|c|c|c|}
    \hline 
  	\textbf{Method}  & mAP     \\ \hline\hline
  	  \textbf{DML} \cite{kordopatis2017b}    & 0.705   \\ \hline
      \textbf{VReL} \cite{feng2018}          & 0.209   \\ \hline
      \textbf{DP} \cite{chou2015}            & 0.621   \\ \hline
      \textbf{TN} \cite{tan2009}             & 0.648   \\ \hline
  \end{tabular}
  \begin{tabular}{|l|c|}
  \hline
      \textbf{Method}  & mAP     \\ \hline\hline
       \textbf{ViSiL}$_f$                           & 0.652   \\ \hline
       \textbf{ViSiL}$_{sym}$                           & \textbf{0.745}   \\ \hline
       \textbf{ViSiL}$_v$                           & 0.710   \\ \hline
  \end{tabular}
  }
  \caption {mAP comparison of three ViSiL setups and four publicly available retrieval methods on ActivityNet based on the reorganization from \cite{feng2018}.}
  \label{tab:map_depth}
\end{table}

\section{Conclusions}

In this paper, we proposed a network that learns to compute similarity between pairs of videos. The key contributions of ViSiL are a) a frame-to-frame similarity computation scheme that captures similarities at regional level and b) a supervised video-to-video similarity computation scheme that analyzes the frame-to-frame similarity matrix to robustly establish high similarities between video segments of the compared videos. Combined, they lead to a video similarity computation method that is accounting for both the fine-grained spatial and temporal aspects of video similarity. The proposed method has been applied to a number of content-based video retrieval problems, where it improved the state-of-art consistently and, in several cases, by a large margin. For future work, we plan to investigate ways of reducing the computational complexity and apply the proposed scheme for the corresponding detection problems (e.g. video copy detection, re-localization). 

\bigskip\noindent\textbf{Acknowledgments:}
This work is supported by the WeVerify H2020 project, partially funded by the EU under contract numbers 825297. The work of Ioannis Patras has been supported by EPSRC under grant No. EP/R026424/1. GKZ also thanks LazyProgrammer for the amazing DL courses.

{\small
\balance
\bibliographystyle{ieee_fullname}
\bibliography{egbib}
}

\newpage

\begin{center}
\Large \textbf{Supplementary materials}
\vspace{1cm}
\end{center}

\renewcommand\thesection{\Alph{section}}

\setcounter{section}{0}

\section{Additional Results}

\subsection{Different similarity calculation functions}

In this section, we compare the impact of different functions, other than CS, on the frame-to-frame (F2F) and video-to-video (V2V) similarity calculation. In general, CS can be considered to be equivalent to a Max-Pooling (MP) function followed by Average-Pooling (AP). A different combination could be the application of two AP functions. Table \ref{tab:diff_pooling} illustrates the results for different combinations of the core similarity functions of the proposed system on FIVR-5K. It is evident that the use of two AP functions for V2V does not work at all. The run with the two AP for F2F and CS for V2V achieves competitive mAP, but still lower than the run with CS in both functions as proposed.

\begin{table}[h]
  \centering
  \begin{tabular}{|c|c|c|c|c|c|}
    \hline
      \textbf{F2F} & \textbf{V2V} &   \textbf{DSVR}   &   \textbf{CSVR}   &   \textbf{ISVR}     \\ \hline\hline
      MP-AP & MP-AP & 0.880   &  0.869   &  0.777  \\ \hline
      AP-AP & MP-AP & 0.769   &  0.748   &  0.682  \\ \hline
      MP-AP & AP-AP & 0.640   &  0.652   &  0.623  \\ \hline
      AP-AP & AP-AP & 0.439   &  0.436   &  0.341  \\ \hline
    \end{tabular}
  \caption{mAP comparison of four pooling combinations for frame-to-frame and video-to-video similarity calculation on FIVR-5K. \textbf{MP} stands for Max-Pooling and \textbf{AP} for Average-Pooling.}
  \label{tab:diff_pooling}
\end{table}

\begin{figure*}[t]
\centering
\subfigure[]{\includegraphics[width=5.5cm]{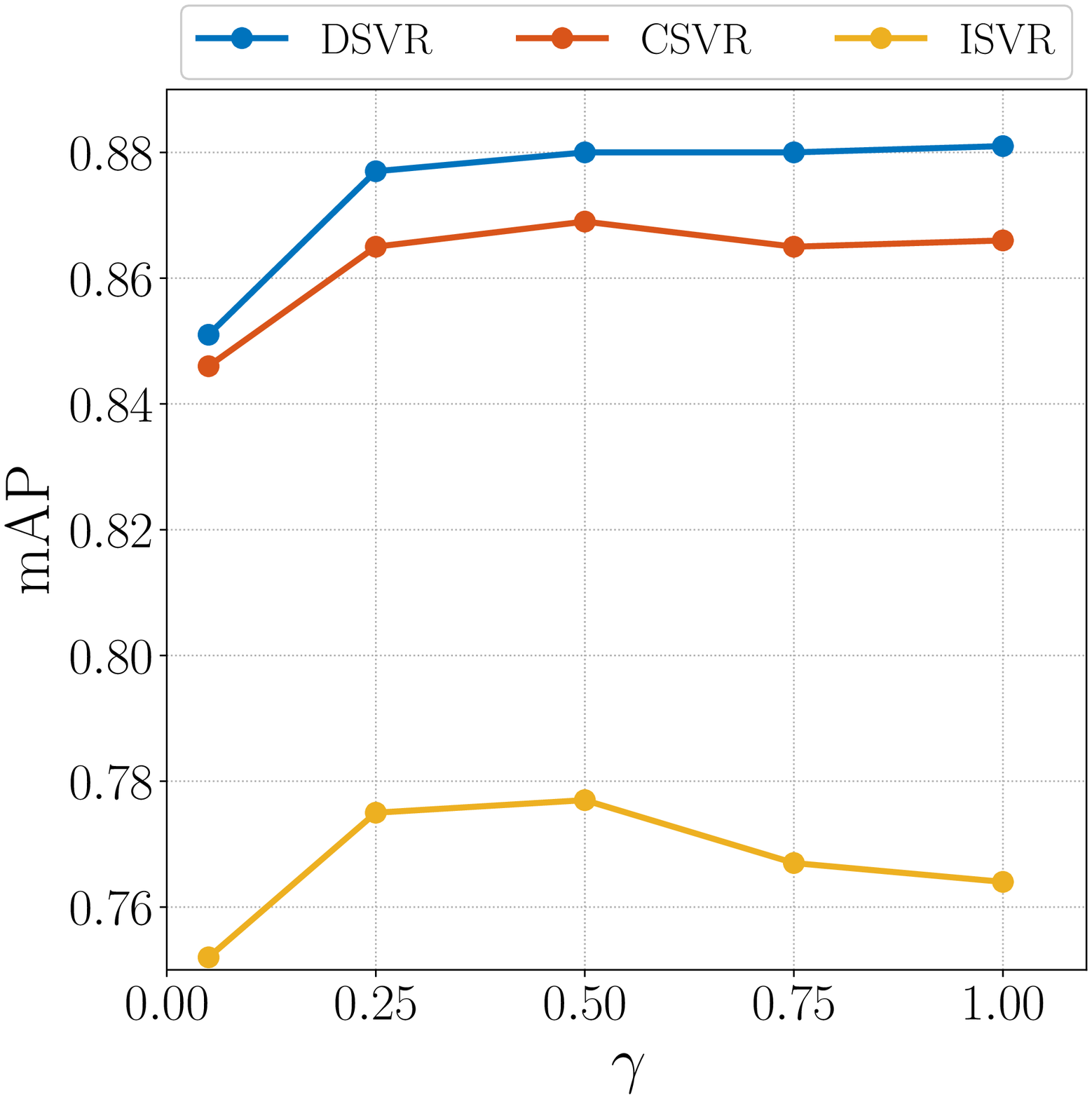}\label{fig:g}}\hspace{0.1cm}
\subfigure[]{\includegraphics[width=5.5cm]{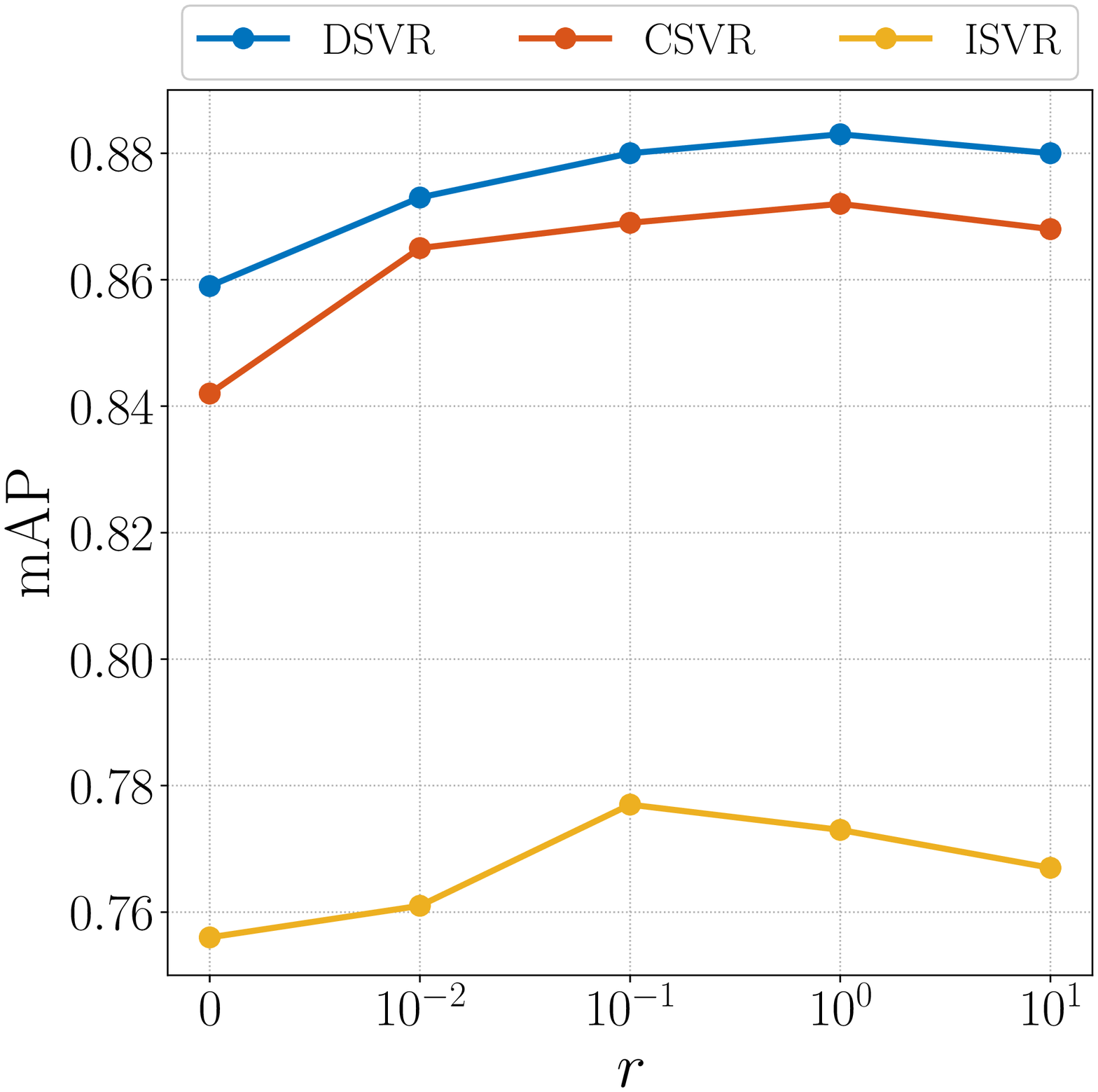}\label{fig:r}}\hspace{0.1cm}
\subfigure[]{\includegraphics[width=5.5cm]{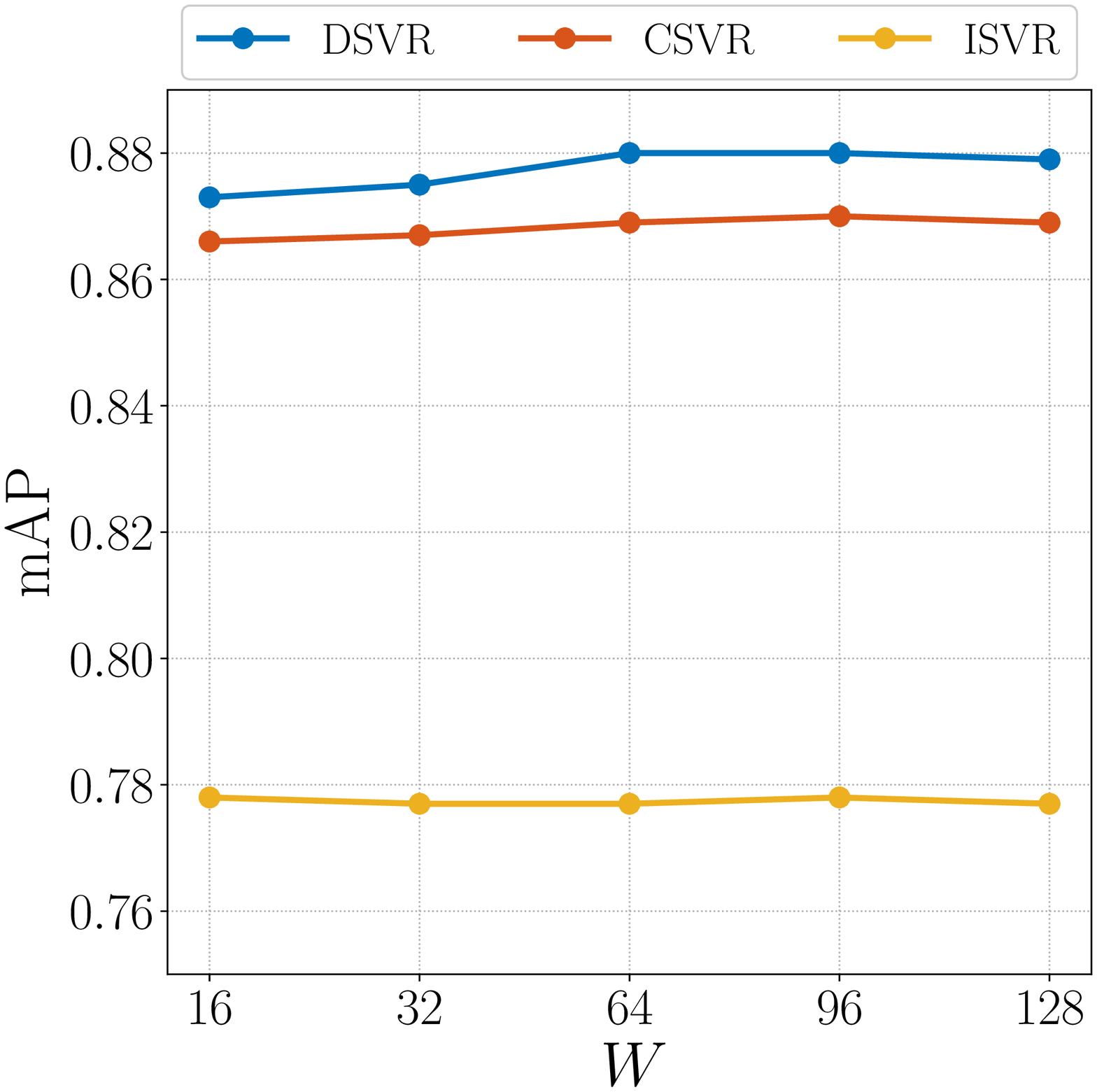}\label{fig:w}}
\caption{Impact of the margin hyperparameter $\gamma$, the regularization parameter $r$ and video snippet size $W$ on the performance of the proposed method on FIVR-5K.}
\label{fig:fivr_pr_curves}
\end{figure*}

\subsection{Impact of hyperparameter values}
In this section, we compare the impact of different values of hyperparameter $\gamma$, $r$ and $W$, on the performance of the proposed system. As default values, we use the values reported in the original paper, i.e. $\gamma = 0.5$, $r=0.1$ and $W=64$, and change one at a time.

We first assess the impact of the margin parameter $\gamma$ on the retrieval performance of the proposed approach. Figure \ref{fig:g} illustrates the performance of the method trained with different margins on the three tasks of FIVR-5K. Regarding the DSVR task, one may notice that that the performance of the model improves as the margin parameter increases. However, this is not the case for the ISVR task. The approach reports high performance (mAP greater than 0.775) for small values of $\gamma$, i.e. within range [0.25, 0.5], but performance drops as $\gamma$ increases.


Additionally, we assess the impact of the regularization parameter $r$ on the retrieval performance of the proposed approach. Figure \ref{fig:r} illustrates the performance of the method trained with different regularization parameters on the three tasks of FIVR-5K. On DSVR and CSVR tasks, the proposed approach achieves the best results for $r=1.0$ with considerable margin from the second best, approximately 0.003 mAP. However, on the ISVR task, the performance significantly dropped in comparison to the default value ($r=0.1$). For values lower than the default, the proposed approach does not report competitive results on any evaluation task.


Finally, we assess the impact of the size of video snippet $W$ on the retrieval performance of the proposed approach. Figure \ref{fig:w} depicts the mAP of the method with different values of $W$ on the three tasks of FIVR-5K dataset. Regarding the DSVR and CSVR tasks, it is evident that the larger the size of video snippets $W$ the better the performance of the proposed methods. The run with $W=96$ yields the best results on both tasks with 0.880 and 0.870 mAP, respectively. However, the system's performance on the ISVR task is independent of the size of video snippets used for training, since all runs report approximately the same mAP.


\subsection{Computational complexity}
In this section, we compare the computational complexity of different setups of the proposed approach. The proposed method can be split in two distinct processes, an offline and an online. The offline process comprises the feature extraction from video frames, whereas the online one the similarity calculation between two videos. 

In Table \ref{tab:computational_time}, we compare the MAC and iMAC runs (cf. Table 2 of the paper) with the ViSiL$_f$ and ViSiL$_v$ in terms of execution time and performance. In that way, we assess the trade-off between the performance gain from the introduction of each component of the method, and the associated computational cost. The average length of videos in FIVR-5K is 103 seconds. All the experiments were executed on a machine with an Intel i7-4770K CPU and a GTX1070 GPU.

For the offline process, all runs need approximately the same time to extract frame features. The use of intermediate convolutional layer does not slow down the feature extraction process, since both MAC and iMAC needs 950 ms for feature extraction. The extraction of regional vectors (ViSiL$_f$) has minor impact on the speed, approximately 1\% increase of the total extraction time. Also, the application of whitening and attention-based weighting does not significantly increases the extraction time; ViSiL$_v$ needs 80 ms more than ViSiL$_f$ per video.

Regarding the online process, the complexity of calculating the frame-to-frame similarity matrix between videos of $M$ frames each, is $O(M^2 N^2)$, where $N$ is the number of regions per frame. This is to be compared to $O(M^2)$ of frame-to-frame methods such as iMAC (where $N=1$). Based on our experiments, the MAC and iMAC runs need less than 2.5 ms to calculate video similarity. The computation of the proposed frame-to-frame similarity matrix increases the execution time by 3.7 ms, which is more than a 150\% increase (comparing iMAC and ViSiL$_f$). Finally, in ViSiL$_v$, the second-stage CNN on the frame-to-frame similarity matrix takes 40\% of the execution time, and further increasing it approximately by 3.5 ms but for a significant performance gain. 

\begin{table}[t]
  \centering
  \scalebox{0.95}{
  \begin{tabular}{|l|}
  \hline
      \multirow{2}{*}{\textbf{Run}} \\ \\ \hline\hline
      \textbf{MAC}     \\ \hline
      \textbf{iMAC}    \\ \hline
      \textbf{ViSiL}$_f$    \\ \hline
      \textbf{ViSiL}$_v$    \\ \hline
  \end{tabular}
  \begin{tabular}{|c|c|}
      \hline
      \multicolumn{2}{|c|}{\textbf{Comp. Time}} \\ \hline
      \textbf{Offline} &  \textbf{Online}      \\ \hline\hline
      0.95s    & 2.0ms  \\ \hline
      0.95s    & 2.3ms  \\ \hline
      0.96s    & 6.0ms  \\ \hline 
      1.04s    & 9.5ms  \\ \hline 
    \end{tabular}
  \begin{tabular}{|c|c|c|}
    \hline
      \multicolumn{3}{|c|}{\textbf{FIVR-5K}} \\ \hline
\textbf{DSVR}   &   \textbf{CSVR}   &   \textbf{ISVR}   \\ \hline\hline
      0.747   &  0.730   &  0.684  \\ \hline
      0.755   &	 0.749   &  0.689  \\ \hline
      0.838   &  0.832   &  0.739  \\ \hline
      0.880   &  0.869   &  0.777   \\ \hline
    \end{tabular}
    }
  \caption{mAP and execution time comparison of four versions of the proposed approach on FIVR-5K. The execution time of the offline process refers to the average feature extraction time per video. The execution time of the online process refers to the average time  for the calculation of video similarity of video pairs.}
  \label{tab:computational_time}
\end{table}

\nobalance

\section{Visual Examples}
This section presents some visual examples of the outputs of the system components. 

\begin{figure*}[t]
\centering
  \begin{tabular}{ccc}
    \includegraphics[width=4.5cm]{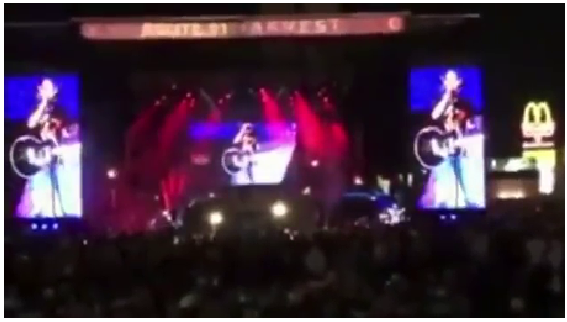} & \includegraphics[width=4.5cm]{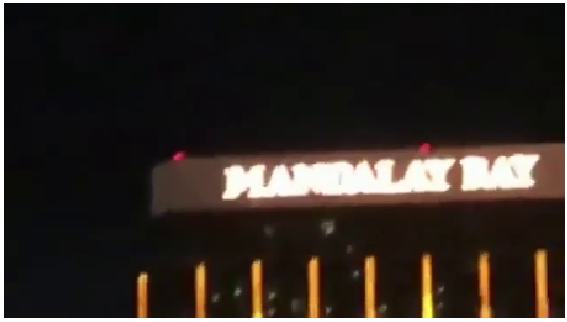} & \includegraphics[width=4.5cm]{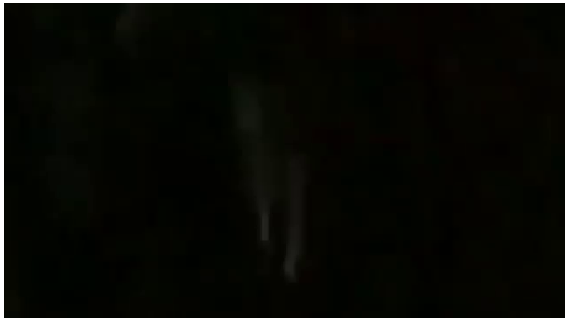} \\
    \includegraphics[width=4.5cm]{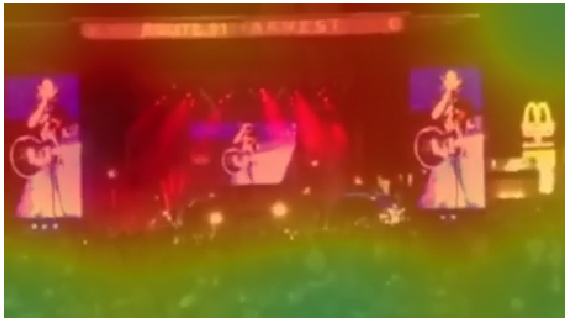} & \includegraphics[width=4.5cm]{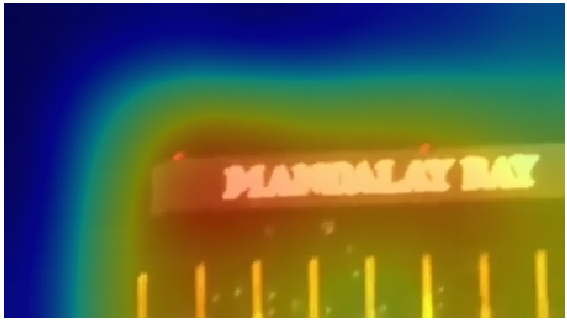} &
    \includegraphics[width=4.5cm]{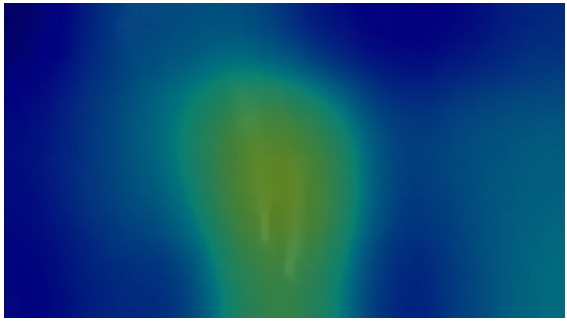}
  \end{tabular}
\caption{Examples of the attention weighting on arbitrary video frames: sampled video frames from the same video (top), attention maps of the corresponding frames (bottom). Red colour indicates high attention weights, whereas blue indicates low ones.}
\vspace{0.1cm}
\label{fig:attention}
\end{figure*}

Figure \ref{fig:attention} illustrates three visual examples of video frames coloured based on the attention weights of their regions vectors. Apparently, the proposed attention mechanism weights the frame regions independently based on their saliency. It assigns high weight values on the information-rich regions (e.g. the concert stage, the Mandalay Bay building); whereas, it assigns low values on regions that contain no meaningful object (e.g. solid dark regions). 

\begin{figure*}[t]
\flushleft
\includegraphics[height=14.5cm]{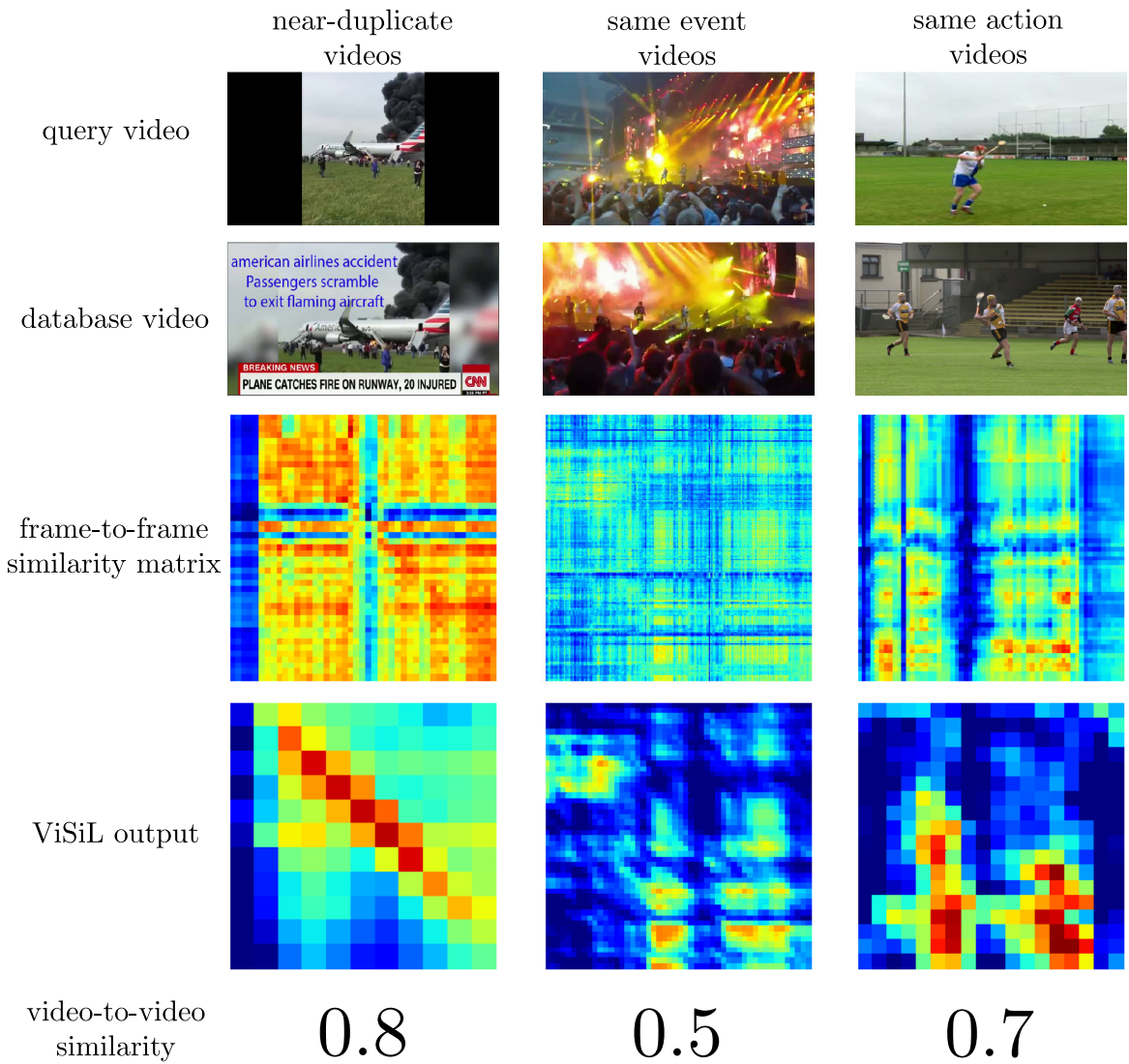}
\vspace{0.2cm}
\caption{Visual examples of the input and output of ViSiL for three different video relation types. Two sampled frames of the compared videos are depicted on top, then the input frame-to-frame similarity matrix and the ViSiL output are displayed, and the final video-to-video similarity is reported. In the similarity matrices, red colour indicates a high similarity score, whereas blue indicates low similarity.}
\label{fig:in_out_examples}
\end{figure*}

Additionally, Figure \ref{fig:in_out_examples} illustrates examples of the input frame-to-frame similarity matrix, the network output and the calculated video similarity of two compared videos for three video categories. The network is able to extract temporal patterns from the input frame-to-frame similarity matrices (e.g. strong diagonals, consistent parts with high similarity) and suppress the noisy (i.e. small inconsistent parts with high similarity values), in order to calculate the final video-to-video similarity precisely. Also, sampled frames from the compared videos are depicted for the better understanding of the different video relation types.

\end{document}